\documentclass[a4paper,fleqn]{cas-dc}

\usepackage[T1]{fontenc}
\usepackage{graphicx}
\usepackage{booktabs}
\usepackage{multirow}
\usepackage{amssymb}
\usepackage{stfloats}

\usepackage{amsmath}
\usepackage{cite}
\usepackage{tikz}
\usepackage{caption}
\usepackage{arydshln}
\usepackage{placeins}
\usepackage{fontawesome5}
\usetikzlibrary{arrows.meta, positioning, fit, calc}
\hypersetup{
    colorlinks=true,
    citecolor=blue,
    linkcolor=blue,
    urlcolor=blue}

\def\tsc#1{\csdef{#1}{\textsc{\lowercase{#1}}\xspace}}
\tsc{WGM}
\tsc{QE}

\begin{document}
\let\WriteBookmarks\relax
\def\floatpagepagefraction{1}
\def\textpagefraction{.001}

\shorttitle{Convergent Performance, Divergent Mechanisms}    

\shortauthors{Diana-Albelda, Kekwick et al.}
\title [mode = title]{Convergent Performance, Divergent Mechanisms in LoRA-Adapted Vision Foundation Models for Brain MRI}



%



\author[1]{Cecilia Diana-Albelda}[orcid=0009-0009-9210-0853]
\cormark[1] 
\fnmark[1] 
\ead{cecilia.diana.albelda@gmail.com}

\author[2]{Fiona Kekwick}[orcid=0009-0001-2391-7378]
\cormark[1] 
\fnmark[1] 
\ead{f.kekwick23@imperial.ac.uk}

\author[2]{Raghav Mehta}

\author[2]{Aleksandra Kulbaka}

\author[3, 4, 5]{Paul M. Matthews}

\author[1]{Jesus Bescos}

\author[1]{\'Alvaro Garc\'ia-Mart\'in}

\author[2, 3]{Wenjia Bai}

\author[2, 6]{Bernhard Kainz}


\affiliation[1]{organization={Department of Electronic Technology and Communications, Universidad Aut\'onoma de Madrid},
            city={Madrid},
            country={Spain}}

\affiliation[2]{organization={Department of Computing, Imperial College London},
            city={London},
            country={United Kingdom}}

\affiliation[3]{organization={Department of Brain Sciences, Imperial College London},
            city={London},
            country={United Kingdom}}

\affiliation[4]{organization={UK Dementia Research Institute Centre, Imperial College London},
            city={London},
            country={United Kingdom}}

\affiliation[5]{organization={Rosalind Franklin Institute},
            city={Didcot},
            country={United Kingdom}}

\affiliation[6]{organization={Department of Artificial Intelligence in Biomedical Engineering, Friedrich-Alexander Universit\"at Erlangen-N\"urnberg},
            city={Erlangen},
            country={Germany}}



\cortext[1]{Corresponding author}
\fntext[1]{These authors contributed equally to this work.}

\begin{abstract}
Adapting Vision Foundation Models (VFMs) to medical imaging via LoRA has become standard practice, yet model selection is guided almost entirely by downstream task metrics, treating backbones with similar performance as interchangeable. We investigate whether equal performance implies equal internal representational shift. We analyse four ViT-B/16 backbones, DINOv3, SigLIP2, MerMED-FM, and BiomedCLIP, pre-trained with different objectives (self-distillation vs.\ contrastive vision-language) and on different data domains (natural vs.\ medical images). We adapt each to brain tumour segmentation and evaluate on six datasets covering both segmentation and Alzheimer's disease classification. Using attention divergence, weight-change analysis, and activation patching, we find that the four models converge to similar segmentation performance (mean Dice 78--81\%) through markedly different internal dynamics, yet diverge in cross-task transfer: segmentation adaptation improves Alzheimer's disease classification for DINOv3 but degrades it for the other three backbones. We reveal a dissociation between modification magnitude and causal importance: the most heavily modified layers are not the causally relevant ones, with early-to-mid layers identified as sufficient for segmentation recovery. Restricting LoRA to these layers matches or approaches full-layer segmentation while reducing classification degradation in the three affected backbones, providing a mechanistically grounded strategy that is more parameter-efficient and more robust than standard full-layer adaptation. Our code is publicly available at \url{https://github.com/fkekwick/Brain_Lora}.
\end{abstract}




\begin{keywords}
Vision Foundation Models \sep LoRA \sep Brain Tumour Segmentation \sep Mechanistic Interpretability \sep Activation Patching \sep Layer-Selective Adaptation 
\end{keywords}

\maketitle

\section{Introduction}
Vision Foundation Models (VFMs), which are pretrained on large-scale image datasets, provide powerful visual feature encoders applicable to diverse downstream computer vision tasks~\cite{menon2026scientific}. Adapting them via Parameter-Efficient Fine-Tuning (PEFT) has become standard practice, with methods like LoRA~\cite{hu2022lora} achieving competitive results across diverse data domains~\cite{veasey2024parameter,asokan2024federated,ahn2026adaptability,bolhassani2025lora-class,zhong2025pg,hu2024lora, he2026unified} at a fraction of full fine-tuning costs. In medical imaging tasks, such as brain tumour segmentation, LoRA-adapted foundation models now rival fully supervised baselines~\cite{diana2026gbt-lorasegsam, dutt2023parameter}. Given the numerous publicly available VFMs backbones~\cite{simeoni2025dinov3, tschannen2025siglip, zhou2026mermed, zhang2023biomedclip}, choosing which backbone best serves the downstream task becomes a key design decision. Currently, model selection is guided almost entirely by downstream task metrics~\cite{zoellin2026benchmarking}; i.e., two backbones that reach the same segmentation Dice score are treated as interchangeable. In this paper, we investigate an intriguing mechanistic question: whether equal performance implies equal internal representation shift, as differences in how models reshape their representations may expose hidden generalisability risks and point to different backbone-selection or fine-tuning strategies.

In model deployment, a single backbone often serves multiple downstream tasks. A model whose representations degrade on one task while improving on another poses a deployment risk that task-specific evaluation cannot detect. Understanding how models adapt, not only how well they perform, is therefore necessary for informed backbone selection. Mechanistic interpretability tools developed for natural language and vision models~\cite{sharkey2025open} offer a principled way to characterise this representational shift and to identify which layers drive task performance.

We study four ViT-B/16 backbones, comparing both pre-training objective (self-distillation vs.\ contrastive vision-language) and source domain (natural vs.\ medical images): DINOv3~\cite{simeoni2025dinov3}, SigLIP2~\cite{tschannen2025siglip}, MerMED-FM~\cite{zhou2026mermed}, and BiomedCLIP~\cite{zhang2023biomedclip}. All four models are adapted via LoRA to brain tumour segmentation on multi-parametric MRI and evaluated across six datasets spanning three segmentation tasks (BraTS-GLI~\cite{baid2021rsna-bratsgli,menze2014multimodal,bakas2017advancing}, BraTS-Africa~\cite{adewole2023brain-africa-brats}, BraTS-PED~\cite{kazerooni2024brain-bratsPED}) and three classification tasks (OASIS-1~\cite{marcus2007open-oasis}, ADNI binary, ADNI multi-class~\cite{petersen2010alzheimer-adni}). We analyse the adaptation process through the lens of four complementary mechanistic tools: attention Jensen-Shannon (JS) divergence~\cite{lin1991divergence}, LoRA weight-change ratios~\cite{hu2022lora}, off-brain attention fraction, and layer-wise activation patching~\cite{zhang2024towards-activation-patch}. Guided by the mechanistic diagnosis, we further propose and validate an early-layer LoRA strategy that restricts adaptation to the layers identified as causally important. 

Our main findings are as follows. \textbf{(i)}~Although the four backbones converge to similar segmentation performance after LoRA adaptation, they diverge substantially in cross-task classification transfer. Mechanistic analysis reveals that segmentation convergence masks fundamentally different representational shift profiles, and medical-domain pre-training confers no consistent segmentation advantage. \textbf{(ii)}~The most heavily modified layers are not the causally important ones: activation patching consistently identifies early-to-mid layers as sufficient to recover full segmentation performance from a frozen encoder, while late-layer adaptation can damage cross-task transfer, as shown by the degraded linear probing classification performance. \textbf{(iii)}~Restricting LoRA to early layers matches or approaches full-layer segmentation while improving classification performance relative to all-layer adaptation in the three backbones whose cross-task transfer degrades, providing a mechanistically grounded strategy that is both more parameter-efficient and more robust than adapting all layers.

Together, these findings suggest that integrating layer-wise interpretability tools with multi-task downstream assessment reveals adaptation risks invisible to standard metrics and motivates practical strategies that would not emerge from task-specific evaluation alone. While our study focuses on brain MRI, the analytical framework can be applied to any attention-based encoder and data domain.

\section{Related Work}

Adapting Vision Foundation Models to medical imaging via PEFT has grown rapidly in recent years. LoRA~\cite{hu2022lora} and its variants have been applied to diverse segmentation tasks~\cite{veasey2024parameter, asokan2024federated}, including brain tumours~\cite{diana2026gbt-lorasegsam,hu2024lora}. A common approach in the literature is adapting a single foundation model per study and evaluating it on diverse datasets within one single downstream task, treating the backbone as a fixed design choice justified by its performance metric. Comparative analyses across backbones remain limited to performance tables without examining how different pre-training paradigms shape the internal adaptation~\cite{ahn2026adaptability,bolhassani2025lora-class, ruffini2025benchmarking}.

Beyond downstream metrics, mechanistic interpretability offers tools to examine how models internally reorganise during adaptation, including attention analysis, probing, and causal interventions such as activation patching~\cite{sharkey2025open,zhang2024towards-activation-patch}. Recent work has begun applying these tools to LoRA-tuned models~\cite{ma2026behind}, using mechanistic analysis to guide adapter placement~\cite{sadanandan2026mechanistically, hayou2026plop}, to decode attention head behaviour under distribution shift~\cite{bahador2025mechanistic}, and to disentangle clinical from class-agnostic features during few-shot adaptation~\cite{rahman2025decoupling}. These studies provide valuable insight into individual models but, by analysing a single backbone in isolation, cannot reveal whether the same adaptation procedure produces different internal dynamics across architectures. Furthermore, the connection between mechanistic diagnosis and a concrete change in adaptation strategy remains largely unexplored in the vision domain: a recent exception in NLP uses causal and geometric analysis to guide LoRA placement across layers~\cite{wang2026behavior}, but interpretability findings in medical imaging are still reported descriptively rather than validated through intervention. Our work addresses both gaps by comparing mechanistic profiles across four backbones and by using the diagnostic findings to motivate and test a layer-selective LoRA strategy.

The question of \textit{which} layers to adapt has also received attention. Some studies dynamically allocate rank budgets across layers~\cite{gu2025lora} based on gradient-derived importance scores~\cite{zhang2023adalora, saket2026aletheia, deng2025ralo}, while domain-specific heuristics select layers using task performance proxies~\cite{ahn2026adaptability} or Fisher information~\cite{park2026fora}. These approaches tend to favour later layers, consistent with the intuition that deeper representations are more task-specific~\cite{wan2026tap, matsoukas2022makes}. Our results show that activation patching identifies early-to-mid layers as causally sufficient for segmentation, and that restricting adaptation to these layers preserves cross-task transferability that full-layer adaptation can compromise. This finding motivates layer selection based on causal importance and multi-task evaluation rather than gradient magnitude.

\section{Method}

\begin{figure*}[!ht]
    \centering
    \includegraphics[width=0.9\linewidth]{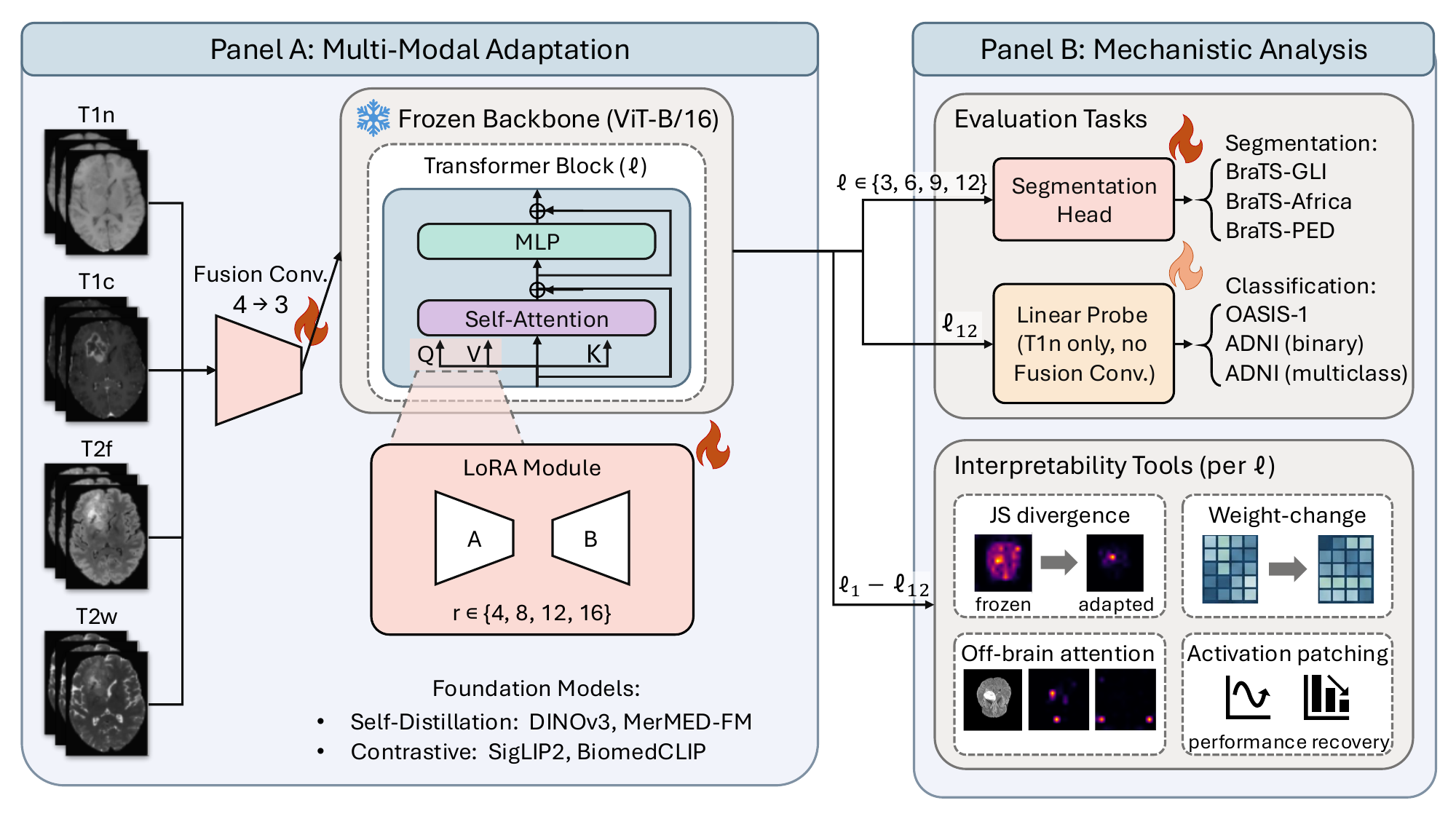}
    \caption{\textbf{Overview of the adaptation and analysis pipeline.} (Panel~A) Four brain MRI modalities (T1n, T1c, T2f, T2w) are fused via a trainable $1\times1$ convolution into a 3-channel input, processed by a frozen ViT-B/16 encoder with trainable LoRA modules applied to the query (Q) and value (V) projections of every transformer block in the default all-layer configuration. Components marked with \textcolor{red}{\faFire} (fusion convolution, LoRA modules, and segmentation head) are jointly trained on BraTS-GLI. (Panel~B) Segmentation is evaluated on three BraTS datasets. For classification, the full encoder (including LoRA weights) remains frozen and only a logistic regression classifier (\textcolor{orange}{\faFire}) is fitted on each classification dataset using mean-pooled features from $\ell_{12}$; the fusion convolution is bypassed (T1n input only). Four mechanistic interpretability tools are applied per layer to analyse the adaptation process: JS divergence, weight-change ratio, off-brain attention fraction, and activation patching.}
    \label{fig:pipeline_overview}
\end{figure*}

\subsection{Adaptation Pipeline}\label{sec:pipeline}

As illustrated in Figure~\ref{fig:pipeline_overview}, we adapt four ViT-B/16 backbones \cite{dosovitskiy2020image} to brain image segmentation task. Four brain MRI modalities (T1n, T1c, T2w, T2f) are fused via a trainable $1\times1$ convolution that maps these four input channels to three, matching the RGB input expected by all encoders. All four modalities carry complementary clinical information~\cite{diana2025multi}, so we retain all of them and let the learned convolution determine the optimal combination. This fusion layer is initialised with uniform weights of $1/4$ with no bias term, equivalent to average pooling, and trained end-to-end independently for each backbone to learn modality-specific contributions.

The ViT-B/16 encoder is kept frozen; adaptation is achieved through LoRA modules injected into each of the 12 transformer blocks. From the adapted encoder, we extract multi-scale representations by tapping four uniformly spaced blocks $\{3, 6, 9, 12\}$ (1-indexed), each taken after the encoder's final layer normalisation and reshaped from the token sequence into a $14\times14$ spatial feature map with 768 channels. The four maps are concatenated channel-wise into a $3072$-channel representation, which a lightweight segmentation head (two $1\times1$ convolutions separated by ReLU, $3072 \rightarrow 256 \rightarrow C$ classes) converts into a dense prediction, upsampled to the ground-truth resolution of $240\times240$ pixels. The LoRA modules and segmentation head are jointly trained with an equal-weight combination of Dice and cross-entropy loss.

We evaluate each backbone in two configurations: \textit{frozen}, where no LoRA modules are applied, the four MRI modalities are mapped to three channels via average pooling, and only the segmentation head is trained; and \textit{adapted}, which includes the trainable fusion convolution and LoRA modules described above. The trainable fusion convolution may contribute to absolute performance differences between configurations; however, all four backbones share the same architecture, so mechanistic comparisons across backbones remain controlled.

\label{sec:lora_config}

We adapt the frozen ViT encoder using LoRA~\cite{hu2022lora} within each transformer self-attention block. For each transformer block~$\ell$, low-rank updates are applied to the query and value projections of the fused $QKV$ linear layer, while the key projection remains unchanged, following the standard LoRA configuration proposed by~\cite{hu2022lora}. For a feature map~$x$, the adapted query~$Q'$ and value projections~$V'$ are given by:
\begin{equation}
Q' = \bigl(W_Q^{(\ell)} + B_Q^{(\ell)} A_Q^{(\ell)}\bigr) x, \qquad V' = \bigl(W_V^{(\ell)} + B_V^{(\ell)} A_V^{(\ell)}\bigr) x,
\end{equation}
where $W_Q^{(\ell)}$, $W_V^{(\ell)}$ are the frozen pretrained weights and $A_Q^{(\ell)}$, $A_V^{(\ell)}$, $B_Q^{(\ell)}$, $B_V^{(\ell)}$ are the trainable low-rank projection matrices, with no dropout. We omit the $\alpha/r$ scaling factor, so the low-rank update $BA$ is applied at full magnitude; since rank sensitivity proved negligible (Section~\ref{sec:results_seg}, Table~\ref{tab:seg-id}), this choice does not affect our conclusions.

\subsection{Mechanistic Analysis} \label{sec:mech_analysis}

We analyse the adaptation process~\cite{kim2026interpreting} through four complementary tools: (1)~attention Jensen-Shannon (JS) divergence~\cite{lin1991divergence}, (2)~LoRA weight-change ratio~\cite{hu2022lora}, (3)~off-brain attention fraction, and (4)~layer-wise activation patching~\cite{zhang2024towards-activation-patch}. JS divergence and weight-change ratio quantify the magnitude of adaptation in output space and parameter space, respectively. As the LoRA update $\Delta W_Q = B_Q A_Q$ modifies the query projection, it directly alters the attention logits $Q \cdot K^\top$, changing the resulting attention distribution. The weight-change ratio measures this modification in parameter space, while JS divergence measures its downstream effect on the attention pattern; together, they capture the same adaptation from complementary viewpoints. Off-brain attention fraction examines the adaptation cost by measuring how much frozen attention must be redirected toward task-relevant regions. Finally, activation patching isolates which layer adaptations causally drive task performance.

\subsubsection{Attention JS Divergence} \label{sec:js_div}

To quantify how LoRA changes each layer's attention pattern, we measure the JS divergence~\cite{lin1991divergence} in units of bits (log base~2) between the frozen~($P_f$) and adapted~($P_a$) attention distributions. For each transformer block~$\ell$, we extract the $N\!\times\!N$ patch-to-patch attention sub-matrix, excluding any special tokens such as the CLS token or DINOv3's register tokens, renormalise each query row to form a valid distribution, and compute
\begin{equation}
  \mathrm{JS}^{(\ell)} = \tfrac{1}{2}\,D_{\mathrm{KL}}(P_f \| M)
                       + \tfrac{1}{2}\,D_{\mathrm{KL}}(P_a \| M),
  \quad M = \tfrac{1}{2}(P_f + P_a),
\end{equation}
with $D_{\mathrm{KL}}$ being Kullback-Leibler divergence~\cite{kullback1951information}. The result is averaged over query patches and attention heads via arithmetic mean. JS divergence is symmetric and bounded in $[0,1]$, enabling stable comparison across layers and models without scale normalisation.

\subsubsection{LoRA Weight Change} \label{sec:weight_change}

We calculate the adaptation magnitude directly in parameter space. For each transformer block~$\ell$, the effective LoRA update~\cite{hu2022lora} to a projection matrix is $\Delta W^{(\ell)} = B^{(\ell)} A^{(\ell)}$, where $A$ and $B$ are the learned low-rank matrices. We materialise the full product $B^{(\ell)} A^{(\ell)}$ and report the relative Frobenius norm $\|\Delta W^{(\ell)}\|_F / \|W^{(\ell)}\|_F$ separately for the query and value projections, where $W^{(\ell)}$ is the corresponding pretrained weight. A high ratio indicates that LoRA substantially changes that layer relative to its original weight parameters.

\subsubsection{Off-Brain Attention Fraction} \label{sec:sink}

To understand why the four encoders differ in adaptation cost, we measure how much frozen attention is directed away from anatomically relevant regions. We obtain a binary brain mask from the input slice resized to $224\times224$ and assign each ViT patch token to $\mathcal{B}$ (brain) or its complement $\bar{\mathcal{B}}$ (off-brain) according to whether more than 5\% of its $16\times16$ pixels are foreground. For each block~$\ell$ and attention head~$h$, let $A^{(\ell,h)}_{q,k}$ be the attention weight from query patch~$q$ to key patch~$k$. We compute the off-brain attention fraction
\begin{equation}
s_\ell = \frac{1}{H\,|\mathcal{Q}|}\sum_{h=1}^{H}\sum_{q \in \mathcal{Q}}\sum_{k \in \bar{\mathcal{B}}} A^{(\ell,h)}_{q,k},
\end{equation}
where $H$ is the number of attention heads and $\mathcal{Q}$ the set of query patches. A high off-brain fraction indicates that the frozen encoder largely ignores the anatomy, implying that LoRA must redirect substantial attention mass before task-relevant features can be encoded.

\subsubsection{Activation Patching} \label{sec:act_patch}

Activation patching isolates which layer adaptations causally drive task performance~\cite{zhang2024towards-activation-patch}. Let~$M_F$ and~$M_A$ denote the task metric obtained with the fully frozen and fully adapted encoder, respectively. For each block~$\ell$, we capture the output of the adapted encoder at that depth and inject it into the frozen forward pass at the same depth, replacing only that block's output. The resulting metric~$M_P^{(\ell)}$ yields a normalised recovery score:
\begin{equation}
R^{(\ell)} = \frac{M_P^{(\ell)} - M_F}{M_A - M_F},
\end{equation}
measuring the fraction of the LoRA-driven gain recovered by providing the adapted representation at depth~$\ell$. Note that $R^{(\ell)}$ does not measure the contribution of block~$\ell$ alone, as the captured activation already incorporates LoRA from all preceding blocks, and all subsequent ones process it with frozen weights. 

An important interaction arises from the multi-scale feature extraction described in Section~\ref{sec:pipeline}. The segmentation head receives features from blocks $\{3, 6, 9, 12\}$ directly, so features tapped before the patching point reach the head as frozen representations, while those tapped after it reflect the propagated adapted activation. $R^{(\ell)}$ is therefore cumulative, quantifying how much of the total improvement is recoverable when the adapted representation is available from depth~$\ell$ onward. A high $R^{(\ell)}$ despite a small weight-change ratio indicates a block whose modest modification is disproportionately important for performance.

\subsection{Layer-Selective LoRA} \label{sec:layer_selective}

To check whether the causal importance patterns identified by activation patching translate into a practical adaptation strategy, we split the 12 transformer blocks at the midpoint, applying LoRA to blocks 1--6 (\textit{early}) or 7--12 (\textit{late}), alongside the default \textit{all} (1--12). The same cut-off is used across all four backbones to test whether a single, model-agnostic configuration is viable rather than optimising the split point per backbone. In each configuration, LoRA modules are injected only into the selected blocks, with non-selected ones retaining their original frozen weights. The segmentation head and fusion convolution are trained from scratch for each configuration to ensure a fair comparison, so performance differences reflect the effect of the LoRA module placement.

\subsection{Linear Probing Classification} \label{sec:linear_probe}

To further assess whether representations learned during segmentation transfer to an unrelated task, we evaluate each backbone on Alzheimer's disease classification via linear probing~\cite{ciernik2026attentive}. Features are extracted from the final transformer block with layer normalisation applied, and the patch token sequence is mean-pooled into a single 768-dimensional vector per slice. The $2D$ representations of all slices belonging to the same subject are then mean-pooled into a single subject-level representation. A logistic regression classifier is fitted on the training set and evaluated on the test set. During feature extraction, the LoRA-adapted segmentation encoder is used as a fixed feature extractor: the adapted weights (including LoRA modules) are loaded but not updated, so the features reflect the post-segmentation-adaptation representations without any further training.

\begin{table*}[t!]
\centering
\fontsize{7.8}{9.5}\selectfont
\setlength{\tabcolsep}{4pt}
\caption{\textbf{In-distribution segmentation performance on BraTS-GLI test set.} Four ViT-B/16 backbones are evaluated in two configurations: \textit{Frozen} (encoder fully frozen, only the segmentation head is trained) and \textit{LoRA} (encoder adapted via LoRA at ranks $r \in \{4, 8, 12, 16\}$, trained jointly with the segmentation head). Dice coefficient (\%) and Hausdorff distance at the 95th percentile (HD95, mm) are reported for three tumour sub-regions: whole tumour (WT = enhancing + necrotic + edema), tumour core (TC = enhancing + necrotic), and enhancing tumour (ENH). All values are mean $\pm$ std across 3 random seeds. Best result per model/metric combination is shown in \textbf{bold}.}
\label{tab:seg-id}
\resizebox{0.98\linewidth}{!}{%
\begin{tabular}{@{}ll cccc ccc@{}}
\toprule
& & \multicolumn{4}{c}{\textbf{Dice (\%) $\uparrow$}} & \multicolumn{3}{c}{\textbf{HD95 (mm) $\downarrow$}} \\
\cmidrule(lr){3-6} \cmidrule(lr){7-9}
\textbf{Backbone} & \textbf{Adaptor} & WT & TC & ENH & Mean & WT & TC & ENH \\
\midrule
\multicolumn{2}{@{}l}{\textit{Self-distillation Learning}} &  &  \\

\multirow{2}{*}{DINOv3 \cite{simeoni2025dinov3}}
    & Frozen & 80.23{\scriptsize$\pm$0.02} & 75.60{\scriptsize$\pm$0.04} & 57.30{\scriptsize$\pm$0.33} & 71.04{\scriptsize$\pm$0.11} & 11.10{\scriptsize$\pm$0.73} & 6.72{\scriptsize$\pm$0.44} & 5.45{\scriptsize$\pm$0.16} \\
     & LoRA ($r=4$)   & 86.91{\scriptsize$\pm$0.04} & 84.20{\scriptsize$\pm$0.11} & 70.51{\scriptsize$\pm$0.37} & 80.54{\scriptsize$\pm$0.13} & 6.02{\scriptsize$\pm$0.33} & 4.74{\scriptsize$\pm$0.42} & 4.28{\scriptsize$\pm$0.10} \\
    & LoRA ($r=8$)   & 87.10{\scriptsize$\pm$0.06} & 84.16{\scriptsize$\pm$0.06} & 71.01{\scriptsize$\pm$0.43} & 80.76{\scriptsize$\pm$0.15} & \textbf{5.67}{\scriptsize$\pm$0.10} & \textbf{4.53}{\scriptsize$\pm$0.09} & 4.19{\scriptsize$\pm$0.16} \\
    & LoRA ($r=12$)   & 87.29{\scriptsize$\pm$0.04} & 84.16{\scriptsize$\pm$0.10} & 71.20{\scriptsize$\pm$0.71} & 80.88{\scriptsize$\pm$0.24} & 5.81{\scriptsize$\pm$0.04} & 5.04{\scriptsize$\pm$0.27} & 4.21{\scriptsize$\pm$0.09} \\
    & LoRA ($r=16$)   & \textbf{87.31}{\scriptsize$\pm$0.06} & \textbf{84.43}{\scriptsize$\pm$0.15} & \textbf{71.87}{\scriptsize$\pm$0.07} & \textbf{81.20}{\scriptsize$\pm$0.06} & 5.75{\scriptsize$\pm$0.24} & 4.62{\scriptsize$\pm$0.25} & \textbf{4.15}{\scriptsize$\pm$0.20} \\
\noalign{\vspace{1.5pt}}\hdashline\noalign{\vspace{1.5pt}}
\multirow{2}{*}{MerMED-FM \cite{zhou2026mermed}}
    & Frozen & 80.18{\scriptsize$\pm$0.13} & 74.73{\scriptsize$\pm$0.23} & 57.16{\scriptsize$\pm$0.31} & 70.69{\scriptsize$\pm$0.14} & 11.11{\scriptsize$\pm$0.13} & 7.81{\scriptsize$\pm$0.84} & 6.40{\scriptsize$\pm$0.49}  \\
    & LoRA ($r=4$)   & 86.12{\scriptsize$\pm$0.18} & 82.90{\scriptsize$\pm$0.21} & 69.02{\scriptsize$\pm$0.18} & 79.35{\scriptsize$\pm$0.18} & 6.46{\scriptsize$\pm$0.17} & \textbf{4.89}{\scriptsize$\pm$0.34} & 4.54{\scriptsize$\pm$0.20} \\
    & LoRA ($r=8$)  & 86.19{\scriptsize$\pm$0.06} & \textbf{82.93}{\scriptsize$\pm$0.18} & 69.20{\scriptsize$\pm$0.78} & \textbf{79.44}{\scriptsize$\pm$0.19} & 6.68{\scriptsize$\pm$0.25} & 4.97{\scriptsize$\pm$0.30} & 4.36{\scriptsize$\pm$0.10}  \\
    & LoRA ($r=12$)   & \textbf{86.30}{\scriptsize$\pm$0.10} & 82.96{\scriptsize$\pm$0.13} & 69.00{\scriptsize$\pm$0.73} & 79.42{\scriptsize$\pm$0.18} & \textbf{6.37}{\scriptsize$\pm$0.18} & 5.11{\scriptsize$\pm$0.64} & 4.46{\scriptsize$\pm$0.15} \\
    & LoRA ($r=16$)   & 86.12{\scriptsize$\pm$0.09} & 82.41{\scriptsize$\pm$0.23} & \textbf{69.39}{\scriptsize$\pm$0.36} & 79.31{\scriptsize$\pm$0.20} & 6.89{\scriptsize$\pm$0.37} & 5.34{\scriptsize$\pm$0.37} & \textbf{4.31}{\scriptsize$\pm$0.12} \\

\noalign{\vspace{1.5pt}}\hline\noalign{\vspace{1.5pt}}
\multicolumn{2}{@{}l}{\textit{Contrastive Learning}} &  & \\

\multirow{2}{*}{SigLIP2 \cite{tschannen2025siglip}}
    & Frozen & 77.81{\scriptsize$\pm$0.33} & 73.53{\scriptsize$\pm$0.15} & 55.13{\scriptsize$\pm$0.90} & 68.82{\scriptsize$\pm$0.32} & 16.62{\scriptsize$\pm$1.79} & 7.57{\scriptsize$\pm$0.54} & 6.10{\scriptsize$\pm$0.14} \\
    & LoRA ($r=4$)   & 86.26{\scriptsize$\pm$0.04} & 82.74{\scriptsize$\pm$0.31} & 69.59{\scriptsize$\pm$0.54} & 79.53{\scriptsize$\pm$0.21} & 6.32{\scriptsize$\pm$0.44} & \textbf{4.89}{\scriptsize$\pm$0.13} & 4.39{\scriptsize$\pm$0.14} \\
    & LoRA ($r=8$)  & 86.54{\scriptsize$\pm$0.03} & \textbf{83.22}{\scriptsize$\pm$0.12} & 69.83{\scriptsize$\pm$0.51} & 79.86{\scriptsize$\pm$0.17} & \textbf{6.20}{\scriptsize$\pm$0.35} & 4.93{\scriptsize$\pm$0.21} & 4.16{\scriptsize$\pm$0.11} \\
    & LoRA ($r=12$)   & \textbf{86.59}{\scriptsize$\pm$0.11} & 82.85{\scriptsize$\pm$0.08} & \textbf{70.78}{\scriptsize$\pm$0.45} & \textbf{80.07}{\scriptsize$\pm$0.16} & 6.48{\scriptsize$\pm$0.36} & 4.93{\scriptsize$\pm$0.30} & \textbf{4.11}{\scriptsize$\pm$0.16} \\
    & LoRA ($r=16$)   & 86.49{\scriptsize$\pm$0.05} & 82.81{\scriptsize$\pm$0.10} & 70.04{\scriptsize$\pm$0.42} & 79.78{\scriptsize$\pm$0.14} & 6.42{\scriptsize$\pm$0.73} & 5.19{\scriptsize$\pm$0.28} & 4.35{\scriptsize$\pm$0.14} \\
\noalign{\vspace{1.5pt}}\hdashline\noalign{\vspace{1.5pt}}
\multirow{2}{*}{BiomedCLIP \cite{zhang2023biomedclip}}
    & Frozen & 76.69{\scriptsize$\pm$0.29} & 70.26{\scriptsize$\pm$0.22} & 52.88{\scriptsize$\pm$0.13} & 66.61{\scriptsize$\pm$0.13} & 14.19{\scriptsize$\pm$1.16} & 9.89{\scriptsize$\pm$0.78} & 8.61{\scriptsize$\pm$0.74}  \\
    & LoRA ($r=4$)   & 84.92{\scriptsize$\pm$0.13} & 80.82{\scriptsize$\pm$0.13} & 67.28{\scriptsize$\pm$0.59} & 77.67{\scriptsize$\pm$0.22} & 7.52{\scriptsize$\pm$0.56} & 5.71{\scriptsize$\pm$0.71} & 4.91{\scriptsize$\pm$0.20} \\
    & LoRA ($r=8$)  & 85.15{\scriptsize$\pm$0.07} & 81.37{\scriptsize$\pm$0.34} & 67.37{\scriptsize$\pm$0.80} & 77.96{\scriptsize$\pm$0.39} & 7.38{\scriptsize$\pm$0.30} & \textbf{5.45}{\scriptsize$\pm$0.40} & 4.81{\scriptsize$\pm$0.36}  \\
    & LoRA ($r=12$)   & \textbf{85.39}{\scriptsize$\pm$0.04} & 81.36{\scriptsize$\pm$0.18} & 67.75{\scriptsize$\pm$0.60} & 78.16{\scriptsize$\pm$0.23} & \textbf{7.19}{\scriptsize$\pm$0.46} & 5.90{\scriptsize$\pm$0.37} & 4.86{\scriptsize$\pm$0.01} \\
    & LoRA ($r=16$)   & 85.35{\scriptsize$\pm$0.11} & \textbf{81.47}{\scriptsize$\pm$0.34} & \textbf{68.01}{\scriptsize$\pm$0.51} & \textbf{78.28}{\scriptsize$\pm$0.27} & 7.63{\scriptsize$\pm$0.46} & 6.90{\scriptsize$\pm$0.55} & \textbf{4.62}{\scriptsize$\pm$0.35} \\
\bottomrule
\end{tabular}}
\end{table*}

\section{Experimental Setup}

\subsection{Backbone Selection} \label{sec:backbone_selection}

We select four backbones analysing the self-supervised objective (self-distillation vs.\ contrastive vision-language) and the training data domain (natural vs.\ medical images). The self-distillation models are DINOv3~\cite{simeoni2025dinov3}, trained on $\sim$1.7 billion natural images, and MerMED-FM~\cite{zhou2026mermed}, trained on $\sim$3.3 million medical images, spanning diverse clinical modalities but excluding MRI. The contrastive models are SigLIP2~\cite{tschannen2025siglip}, trained on $\sim$10 billion natural image-text pairs, and BiomedCLIP~\cite{zhang2023biomedclip}, trained on $\sim$15 million image-text pairs from the biomedical literature. All four share the ViT-B/16 architecture (with 12 layers), isolating the effect of pre-training configuration from architectural differences. We note that pre-training dataset scale differs by three orders of magnitude between natural-domain and medical-domain models; this confound is inherent to the available foundation models and should be considered when interpreting domain-related comparisons.

\subsection{Datasets} \label{sec:datasets}

For glioma tumour segmentation, models are adapted on the BraTS-GLI~\cite{baid2021rsna-bratsgli,bakas2017advancing, menze2014multimodal} adult glioma dataset using all four MRI modalities (T1n, T1c, T2w, T2f) as input. Segmentation is evaluated at the 2D slice level on three nested tumour sub-regions: whole tumour (WT), tumour core (TC, combining enhancing and necrotic cells), and enhancing tumour (ENH, active infiltrating cells).
The dataset comprises $153{,}105$ $2$D slices from $1{,}251$ patients, with a patient-stratified 70/15/15 train/validation/test split. \\ Out-of-distribution segmentation generalisation is evaluated on BraTS-Africa~\cite{adewole2023brain-africa-brats} ($7{,}462$ slices from $60$ patients from Sub-Saharan African populations with glioma) and BraTS-PED~\cite{kazerooni2024brain-bratsPED} ($11{,}923$ slices from $99$ pediatric patients with glioma), without any retraining. For the classification tasks, we use the Open Access Series of Imaging Studies (OASIS-1)~\cite{marcus2007open-oasis} ($39{,}438$ $2$D slices from $416$ subjects) and the Alzheimer's Disease Neuroimaging Initiative (ADNI) datasets~\cite{petersen2010alzheimer-adni} ($174{,}458$ slices from $1{,}494$ subjects), taking the first T1n scan per subject in case of a longitudinal study. We use the default L2 regularisation strength ($C=1$) without tuning, so no validation set is used; each cross-validation fold uses a subject-level 70/30 train/test split (Section~\ref{sec:clasif_impl}). On OASIS-1, we perform binary Alzheimer's disease classification, and on ADNI both binary (CN vs.\ AD) and 3-class (CN vs.\ MCI vs.\ AD) classification, where CN is the label for a scan from a Cognitively Normal study participant, MCI denotes Mild Cognitive Impairment, and AD represents the presence of Alzheimer's Disease. 

\subsection{Preprocessing}
We applied volume-wise z-score normalisation, followed by $2D$ slicing. For all datasets, the selected $2D$ slices were obtained by discarding slices with less than $5\%$ non-zero pixels from the original $3D$ volumes. We clipped each slice to $[-3,3]$, rescaled to $[0,1]$, and resized to $224\times224$. The slices were normalised following each backbone's original pre-training protocol (ImageNet statistics for DINOv3 and MerMED-FM, backbone-specific statistics for SigLIP2, and OpenAI CLIP statistics for BiomedCLIP). 

\subsection{LoRA Adaptation Implementation Details}
All trainable components (LoRA modules, fusion convolution, and segmentation head) are jointly trained for 50 epochs using the AdamW optimiser~\cite{loshchilov2017decoupled} with a batch size of 32 and weight decay of $0.05$. We use a learning rate of $10^{-4}$ for the LoRA modules and $10^{-3}$ for the fusion convolution and segmentation head. Each experiment is run on a single NVIDIA A100 GPU and repeated with 3 random seeds; segmentation results report mean $\pm$ standard deviation across seeds, while classification results are additionally averaged across cross-validation folds.

\subsection{Linear Probing Classification Implementation} \label{sec:clasif_impl}
In this task, only a logistic regression classifier is trained on the features extracted from each backbone, using L2-regularised logistic regression ($C=1$) with class-balanced weights and the L-BFGS solver~\cite{pedregosa2011scikit}, with a maximum of 2,000 iterations. Since OASIS-1 and ADNI provide only single-modality T1 scans, the fusion convolution is not used during feature extraction; instead, the input channel is replicated to three channels and normalised with each backbone's specific preprocessing. The logistic regression classifier uses a 3-fold Monte Carlo cross-validation evaluation. The results are averaged across folds and across encoder training seeds for the LoRA adaptations. The L-BFGS solver is deterministic, so no seeds were varied for the probe.

\section{Results}

\begin{table*}[!ht]
\centering
\fontsize{7.8}{9.5}\selectfont
\setlength{\tabcolsep}{4pt}
\caption{\textbf{Out-of-distribution segmentation performance (Dice \%).} All models are trained on BraTS-GLI and evaluated on BraTS-Africa (Sub-Saharan African adult glioma) and BraTS-PED (pediatric glioma) without retraining. Dice coefficient (\%) is reported for three tumour sub-regions: whole tumour (WT = enhancing + necrotic + edema), tumour core (TC = enhancing + necrotic), and enhancing tumour (ENH). Only $r = 8$ is reported. All values are mean $\pm$ std across 3 random seeds. Best result per column in \textbf{bold}.}
\label{tab:seg-ood}
\resizebox{0.9\linewidth}{!}{%
\begin{tabular}{@{}ll ccc ccc@{}}
\toprule
& & \multicolumn{3}{c}{\textbf{BraTS-Africa}} & \multicolumn{3}{c}{\textbf{BraTS-PED}} \\
\cmidrule(lr){3-5} \cmidrule(lr){6-8}
\textbf{Backbone} & \textbf{Adaptor} & WT & TC & ENH & WT & TC & ENH \\
\midrule
\multicolumn{2}{@{}l}{\textit{Self-distillation Learning}} &  &  \\
\multirow{2}{*}{DINOv3 \cite{simeoni2025dinov3}}
    & Frozen & 69.70{\scriptsize$\pm$2.21} & 58.50{\scriptsize$\pm$0.07} & 41.55{\scriptsize$\pm$0.79} & 55.44{\scriptsize$\pm$0.42} & 35.23{\scriptsize$\pm$1.43} & 29.49{\scriptsize$\pm$1.36} \\
    & LoRA ($r=8$)  & \textbf{85.55}{\scriptsize$\pm$0.30} & \textbf{73.72}{\scriptsize$\pm$0.33} & \textbf{66.94}{\scriptsize$\pm$0.54} & \textbf{79.81}{\scriptsize$\pm$0.79} & \textbf{47.33}{\scriptsize$\pm$1.88} & 41.88{\scriptsize$\pm$0.63} \\
\noalign{\vspace{1.5pt}}\hdashline\noalign{\vspace{1.5pt}}
\multirow{2}{*}{MerMED-FM \cite{zhou2026mermed}}
    & Frozen & 76.19{\scriptsize$\pm$0.48} & 56.39{\scriptsize$\pm$0.20} & 41.85{\scriptsize$\pm$0.79} & 57.13{\scriptsize$\pm$1.44} & 33.70{\scriptsize$\pm$0.90} & 28.15{\scriptsize$\pm$0.38}  \\
    & LoRA ($r=8$)   & 83.73{\scriptsize$\pm$0.29} & 72.85{\scriptsize$\pm$0.30} & 64.77{\scriptsize$\pm$0.08} & 77.88{\scriptsize$\pm$0.69} & 45.72{\scriptsize$\pm$0.46} & 40.36{\scriptsize$\pm$1.41} \\
    
\noalign{\vspace{1.5pt}}\hline\noalign{\vspace{1.5pt}}
\multicolumn{2}{@{}l}{\textit{Contrastive Learning}} &  &  \\

\multirow{2}{*}{SigLIP2 \cite{tschannen2025siglip}}
    & Frozen & 69.49{\scriptsize$\pm$0.54} & 56.67{\scriptsize$\pm$0.55} & 41.58{\scriptsize$\pm$0.94} & 53.13{\scriptsize$\pm$0.85} & 35.27{\scriptsize$\pm$1.33} & 27.16{\scriptsize$\pm$1.94} \\
    & LoRA ($r=8$)  & 84.10{\scriptsize$\pm$0.55} & 72.51{\scriptsize$\pm$0.58} & 65.29{\scriptsize$\pm$0.64} & 79.24{\scriptsize$\pm$0.33} & 46.67{\scriptsize$\pm$0.85} & \textbf{41.95}{\scriptsize$\pm$2.26} \\
\noalign{\vspace{1.5pt}}\hdashline\noalign{\vspace{1.5pt}}
\multirow{2}{*}{BiomedCLIP \cite{zhang2023biomedclip}}
    & Frozen & 68.19{\scriptsize$\pm$2.25} & 51.75{\scriptsize$\pm$0.50} & 38.10{\scriptsize$\pm$0.58} & 46.77{\scriptsize$\pm$0.93} & 29.17{\scriptsize$\pm$0.95} & 24.66{\scriptsize$\pm$0.94}  \\
    & LoRA ($r=8$)   & 82.34{\scriptsize$\pm$0.16} & 69.92{\scriptsize$\pm$0.33} & 62.05{\scriptsize$\pm$0.48} & 75.56{\scriptsize$\pm$0.19} & 42.91{\scriptsize$\pm$0.25} & 37.97{\scriptsize$\pm$0.17} \\
\bottomrule
\end{tabular}}
\end{table*}

\begin{figure*}[H]
  \centering
  \includegraphics[width=0.98\textwidth]{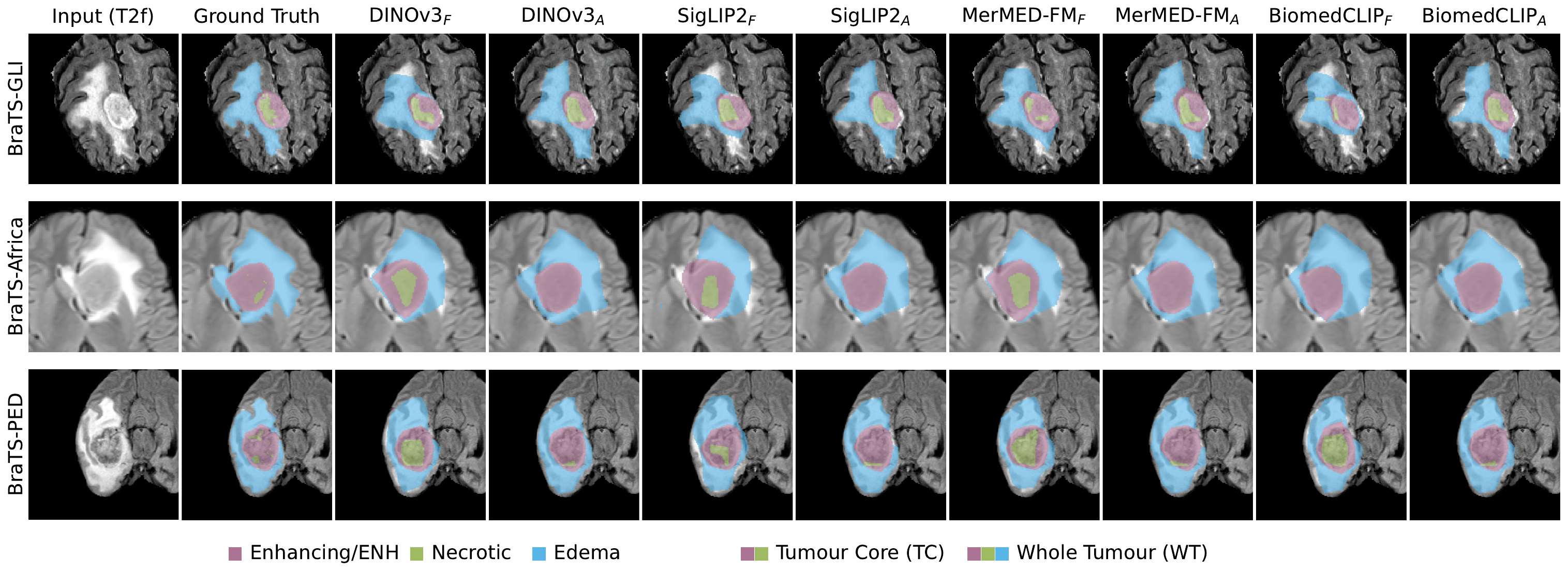}
  \caption{\textbf{Qualitative segmentation results for a representative axial slice from each dataset.} Tumour sub-regions (enhancing, necrotic core, peritumoural edema) are overlaid on the T2f input. Each row corresponds to one dataset (BraTS-GLI, BraTS-Africa, BraTS-PED). For each backbone, the frozen ($F$) and LoRA-adapted ($A$) predictions are shown side by side. LoRA adaptation visibly refines region boundaries across all four backbones.}
  \label{fig:qualitative_seg}
\end{figure*}

\subsection{Segmentation Performance}\label{sec:results_seg}

Table~\ref{tab:seg-id} reports in-distribution segmentation on BraTS-GLI. Across all four backbones, LoRA adaptation yields substantial improvements over the frozen baseline, raising mean Dice by $\sim$8--11\% points regardless of the backbone. After adaptation, the four models converge to a narrow performance range (mean Dice $\sim$78--81\% at $r\!=\!8$), despite their different pre-training configurations. DINOv3 achieves the highest and BiomedCLIP the lowest mean Dice in both the frozen and adapted settings, while SigLIP2 and MerMED-FM lie in between: frozen MerMED-FM outperforms frozen SigLIP2, and after adaptation the two differ by less than 0.5 points. Pre-training domain alone therefore does not determine the ranking. The rank ablation reveals negligible sensitivity: for each backbone, the difference between the best and worst rank is below $\sim$1\% in mean Dice, and the differences between ranks fall within the standard deviation across seeds, confirming that rank tuning is unnecessary for this task. We therefore fix $r = 8$ for all subsequent experiments, following the default configuration in~\cite{hu2022lora}.

Table~\ref{tab:seg-ood} extends the evaluation to out-of-distribution settings. LoRA adaptation improves mean Dice by 15-19\% on both BraTS-Africa and BraTS-PED across all four backbones, suggesting that adaptation captures domain-general brain MRI representations rather than overfitting to in-distribution features. The relative ordering of backbones is preserved, with DINOv3 achieving the highest Dice and BiomedCLIP the lowest. Figure~\ref{fig:qualitative_seg} provides qualitative examples across the three segmentation datasets, illustrating how LoRA sharpens region boundaries for all four encoders. 

\begin{table*}[b]
\centering
\caption{\textbf{Alzheimer's disease classification via linear probing.} Each backbone is evaluated in two configurations: \textit{Frozen} (original pre-trained encoder) and \textit{LoRA} ($r = 8$, adapted for segmentation on BraTS-GLI). Features are mean-pooled from the final transformer block and aggregated per subject; a logistic regression classifier is fitted on the training set. AUC and balanced accuracy (BAcc) are reported on OASIS-1 (binary AD classification) and ADNI (binary and 3-class). All values are mean $\pm$ std across 3 folds for the frozen models, and 3 folds and 3 seeds for the LoRA trained models. Best result per column in \textbf{bold}.}
\small
\setlength{\tabcolsep}{3pt}
\resizebox{0.9\linewidth}{!}{%
\begin{tabular}{ll cc cc cc}
\toprule
\multirow{2}{*}{\textbf{Backbone}} & \multirow{2}{*}{\textbf{Adaptor}} & \multicolumn{2}{c}{\textbf{OASIS-1 (binary)}} & \multicolumn{2}{c}{\textbf{ADNI (binary)}} & \multicolumn{2}{c}{\textbf{ADNI (3-class)}} \\
\cmidrule(lr){3-4} \cmidrule(lr){5-6} \cmidrule(lr){7-8}
 &  & AUC & BAcc (\%) & AUC & BAcc (\%) & AUC & BAcc (\%) \\
\midrule
\multicolumn{2}{@{}l}{\textit{Self-distillation Learning}} &  &  \\

DINOv3 \cite{simeoni2025dinov3} & Frozen & 0.839{\scriptsize$\pm$0.026} & 71.40{\scriptsize$\pm$2.43} & 0.747{\scriptsize$\pm$0.026} & 69.03{\scriptsize$\pm$1.18} & 0.606{\scriptsize$\pm$0.023} & 43.47{\scriptsize$\pm$2.93}  \\
 & LoRA ($r=8$) & 0.868{\scriptsize$\pm$0.029} & 75.44{\scriptsize$\pm$3.13} & 0.799{\scriptsize$\pm$0.017} & 72.40{\scriptsize$\pm$1.03} & 0.637{\scriptsize$\pm$0.012} & 45.94{\scriptsize$\pm$0.98}  \\
\noalign{\vspace{1.5pt}}\hdashline\noalign{\vspace{1.5pt}}
\multirow{2}{*}{MerMED-FM \cite{zhou2026mermed}}
     & Frozen & \textbf{0.878}{\scriptsize$\pm$0.020} & \textbf{78.68}{\scriptsize$\pm$1.22} & 0.835{\scriptsize$\pm$0.038} & 74.02{\scriptsize$\pm$4.10} & \textbf{0.654}{\scriptsize$\pm$0.008} & \textbf{47.88}{\scriptsize$\pm$2.71}  \\
    & LoRA ($r=8$)   & 0.830{\scriptsize$\pm$0.019} & 72.77{\scriptsize$\pm$1.12} & 0.772{\scriptsize$\pm$0.027} & 71.17{\scriptsize$\pm$1.43} & 0.603{\scriptsize$\pm$0.013} & 42.79{\scriptsize$\pm$2.55} \\
 
 \noalign{\vspace{1.5pt}}\hline\noalign{\vspace{1.5pt}}
 \multicolumn{2}{@{}l}{\textit{Contrastive Learning}} &  &  \\

SigLIP2 \cite{tschannen2025siglip} & Frozen & 0.850{\scriptsize$\pm$0.025} & 74.30{\scriptsize$\pm$3.48} & 0.811{\scriptsize$\pm$0.018} & 73.52{\scriptsize$\pm$1.86} & 0.617{\scriptsize$\pm$0.003} & 41.61{\scriptsize$\pm$0.35}  \\
 & LoRA ($r=8$) & 0.816{\scriptsize$\pm$0.022} & 72.02{\scriptsize$\pm$1.80} & 0.746{\scriptsize$\pm$0.009} & 68.21{\scriptsize$\pm$0.80} & 0.597{\scriptsize$\pm$0.007} & 42.36{\scriptsize$\pm$0.85}  \\
    \noalign{\vspace{1.5pt}}\hdashline\noalign{\vspace{1.5pt}}
\multirow{2}{*}{BiomedCLIP \cite{zhang2023biomedclip}}
     & Frozen & 0.858{\scriptsize$\pm$0.019} & 75.58{\scriptsize$\pm$2.97} & 0.772{\scriptsize$\pm$0.027} & 70.03{\scriptsize$\pm$4.20} & 0.633{\scriptsize$\pm$0.018} & 46.29{\scriptsize$\pm$1.26}  \\
    & LoRA ($r=8$)   & 0.855{\scriptsize$\pm$0.021} & 74.79{\scriptsize$\pm$4.85} & 0.752{\scriptsize$\pm$0.013} & 68.46{\scriptsize$\pm$1.18} & 0.607{\scriptsize$\pm$0.014} & 42.85{\scriptsize$\pm$1.63} \\
\bottomrule
\end{tabular}}
\label{tab:crosstask_probe}
\end{table*}

\begin{figure*}[t]
    \centering
    \includegraphics[width=0.98\linewidth]{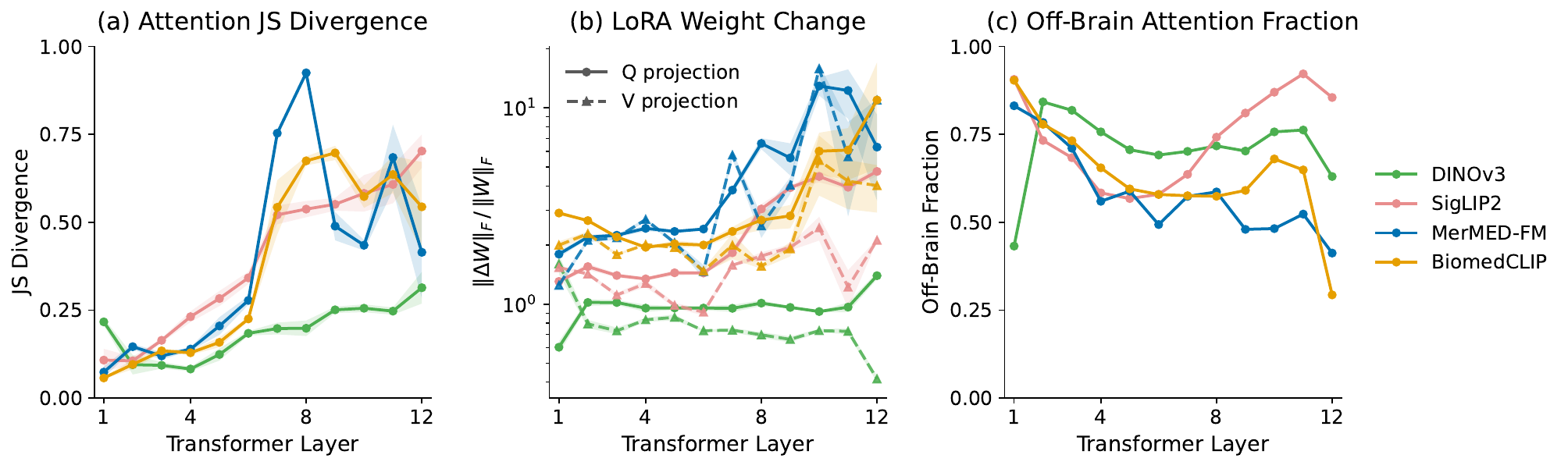}
    \caption{\textbf{Per-layer mechanistic analysis of LoRA adaptation.} (a)~Attention Jensen-Shannon divergence between frozen and adapted attention distributions. (b)~LoRA weight-change ratio ($\|\Delta W\|_F / \|W\|_F$), shown separately for query (Q, solid) and value (V, dashed) projections on a log scale. (c)~Frozen off-brain attention fraction, measuring the fraction of attention mass directed to off-brain patches before adaptation. All metrics are computed on the BraTS-GLI test set and averaged across patients and MRI slices. Shaded bands show $\pm 1$ standard deviation across 3 seeds (absent in (c), which uses only pre-trained weights).}
    \label{fig:mechanistic_panels}
\end{figure*}

 \subsection{Classification Cross-task  Transfer}\label{sec:results_class}

Table~\ref{tab:crosstask_probe} evaluates whether representations learned during segmentation adaptation transfer to Alzheimer's disease classification on OASIS-1 and ADNI. The four backbones exhibit divergent behaviours. DINOv3 is the only model for which LoRA adaptation benefits classification: AUC improves consistently across all three settings. For the remaining three backbones, adaptation harms classification to varying degrees~\cite{ahmad2026catastrophic}: MerMED-FM shows the largest drops (e.g.\ 0.878 to 0.830 AUC on OASIS-1), SigLIP2 degrades consistently across all settings, and BiomedCLIP remains approximately neutral on OASIS-1 but drops on ADNI. Notably, frozen MerMED-FM achieves the highest classification performance across all three settings, yet loses this advantage entirely after adaptation. This divergence in classification performance is not predicted by the segmentation metrics, where all four models perform similarly, carrying direct implications for multi-task deployment.

\begin{figure*}[!t]
    \centering
    \includegraphics[width=1\linewidth]{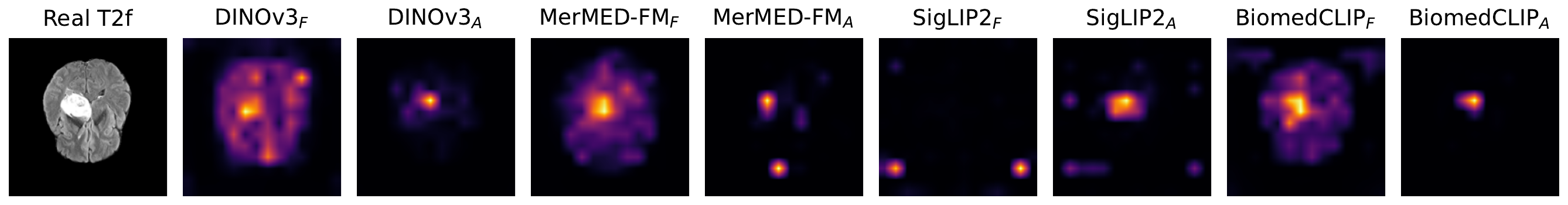}
    \caption{\textbf{Last-layer attention maps for a BraTS-GLI patient.} Each column shows one backbone in its frozen ($F$) and LoRA-adapted ($A$) configuration. Maps show block-12 attention from the CLS token to the patch tokens, averaged over heads; since SigLIP2 has no CLS token, its map shows the attention each patch receives, averaged over all patch queries. Each panel uses its own colour scale, so only spatial patterns are comparable. The frozen medical-domain models (MerMED-FM, BiomedCLIP) attend to anatomical structures, whereas the frozen natural-domain models (DINOv3, SigLIP2) distribute attention more diffusely. After adaptation, all four models redirect attention toward the tumour.}
    \label{fig:att_maps}
\end{figure*}

\subsection{Per-Layer Mechanistic Analysis}\label{sec:results_mech}

Figure~\ref{fig:mechanistic_panels} reveals that the convergent segmentation performance reported above conceals markedly different internal adaptation dynamics across the four backbones.

Figure~\ref{fig:mechanistic_panels}~(a) shows that all four models exhibit increasing attention JS divergence toward later layers, but with different magnitudes: DINOv3 remains the most stable (below 0.3 throughout), while SigLIP2 escalates steeply to ${\sim}0.75$ at layer~12. MerMED-FM and BiomedCLIP fall between these extremes. This suggests that DINOv3 requires the least representational reorganisation to solve the segmentation task, consistent with its superior in-distribution and out-of-distribution performance (Tables~\ref{tab:seg-id} and~\ref{tab:seg-ood}). Conversely, models that undergo larger attention shifts risk overwriting representations useful for other tasks, which aligns with the classification degradation observed in Section~\ref{sec:results_class}.

Figure~\ref{fig:mechanistic_panels}~(b) shows a different grouping in parameter space: the two medical-domain models (MerMED-FM, BiomedCLIP) exhibit the highest weight-change ratios in later layers, while DINOv3 maintains the most uniform and lowest ratios across all layers. Notably, the grouping here follows pre-training domain rather than objective: both medical-domain models require disproportionately large late-layer updates despite their prior exposure to clinical images, suggesting that domain-relevant pre-training does not reduce the magnitude of adaptation needed for a specific downstream task.

Figure~\ref{fig:mechanistic_panels}~(c) provides context for the adaptation cost of each backbone. In the frozen encoders, the two medical-domain models (MerMED-FM, BiomedCLIP) allocate a lower fraction of their attention to off-brain patches than the natural-domain models (DINOv3, SigLIP2), particularly in middle-to-late layers, consistent with prior exposure to medical images during pre-training. Figure~\ref{fig:att_maps} illustrates this qualitatively: the frozen medical-domain models attend to anatomical structures, whereas the frozen natural-domain models distribute attention more diffusely across background regions. After LoRA, all four models redirect attention toward the tumour, but the amount of redistribution required varies across backbones. Taken together, panels (b) and (c) present an apparent paradox: the medical-domain models already attend to anatomically relevant regions yet require the largest parameter updates. This suggests that attending to the right regions is not sufficient~\cite{raghu2021vision}, as the quality of the underlying representations, rather than attention alignment alone, determines how efficiently a backbone can be adapted.

\begin{figure*}[t]
\centering
\includegraphics[width=0.9\linewidth]{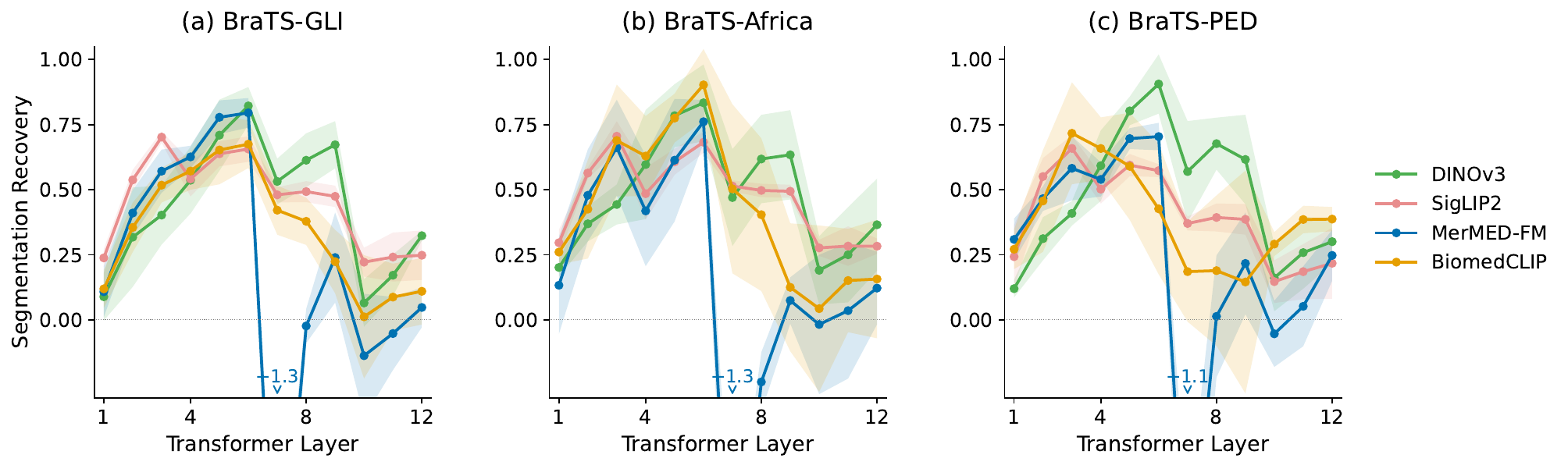} 
\includegraphics[width=0.9\linewidth]{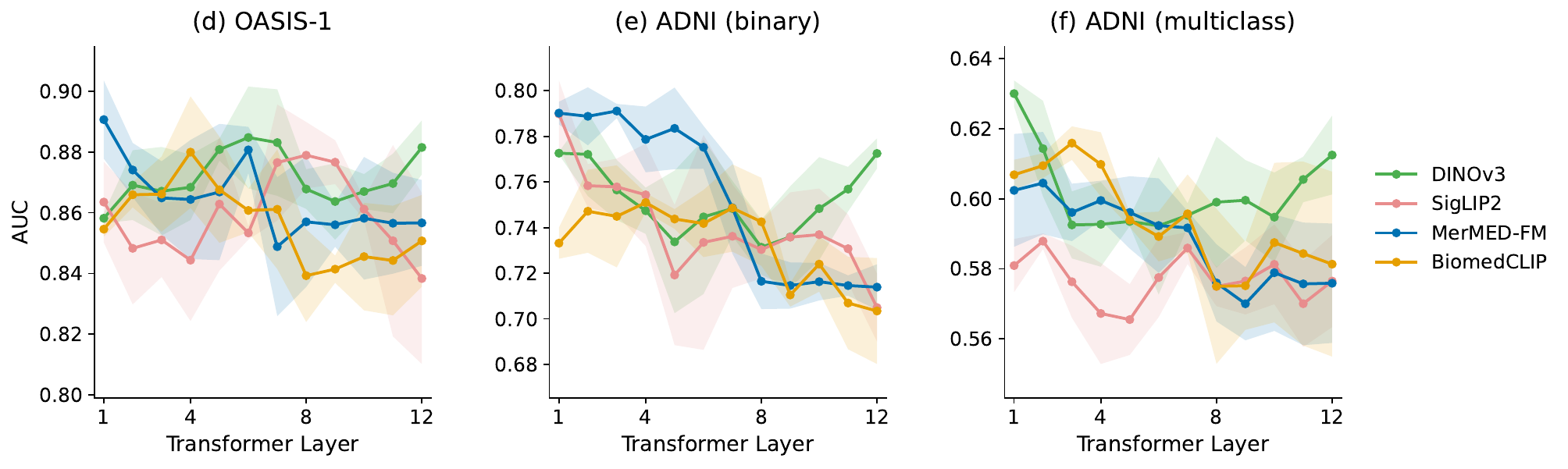} 
\caption{\textbf{Activation patching results across segmentation and classification datasets.} For each block~$\ell$, the adapted activation is injected into an otherwise frozen forward pass and the resulting task metric is recorded. Note that $R^{(\ell)}$ is cumulative: the patched activation at block~$\ell$ already incorporates LoRA from all preceding blocks. Top row (a--c): normalised segmentation recovery $R^{(\ell)}$ on three BraTS datasets; values above zero indicate improvement over the frozen baseline, and values below zero indicate degradation. Bottom row (d--f): absolute classification AUC on three Alzheimer's disease benchmarks (not normalised, since the frozen-to-adapted gap varies in sign across backbones); the logistic regression probe is refit on the patched features for each block. All metrics are averaged across patients and MRI slices. Shaded bands show $\pm 1$ standard deviation across 3 seeds.}
\label{fig:activation_patching}
\end{figure*}

\subsection{Activation Patching Recovery Analysis}\label{sec:results_patching}

Figure~\ref{fig:activation_patching} presents activation patching results across all six evaluation datasets. For each layer~$\ell$, the adapted encoder's output at that depth is injected into an otherwise frozen forward pass; the resulting task metric indicates how much of the full adaptation benefit is recovered when adapted representations are available from depth~$\ell$ onward. In segmentation (panels a-c), a consistent pattern emerges: across all four backbones and three datasets, performance recovery is concentrated in the first half of the encoder (layers 1--6), with recovery declining in later layers. This stands in contrast with the mechanistic profiles in Figure~\ref{fig:mechanistic_panels}, where the largest JS divergence and weight-change ratios concentrate in layers 8--12, revealing a dissociation between modification magnitude and causal importance. A notable case is MerMED-FM, which exhibits negative recovery specifically at layer~7 across all three segmentation datasets and all seeds (9/9 conditions), meaning that injecting the adapted representation at this depth produces worse performance than the fully frozen encoder. This layer coincides with a $\times 4$ jump in the value-projection weight change ($\|\Delta W_V\|$) relative to the adjacent layer. This suggests that the large value-projection rewrite produces activations incompatible with the subsequent frozen blocks.

Classification (panels d-f) shows more heterogeneous patterns across backbones, but the overall trend is consistent: later layers tend to reduce rather than improve AUC for all four models. Taken together, the segmentation and classification patching results consistently point to the first half of the encoder as the causally important region, suggesting that restricting adaptation to early layers could preserve segmentation performance while mitigating classification degradation.

\begin{figure*}[t]
    \centering
    \includegraphics[width=1.0\linewidth]{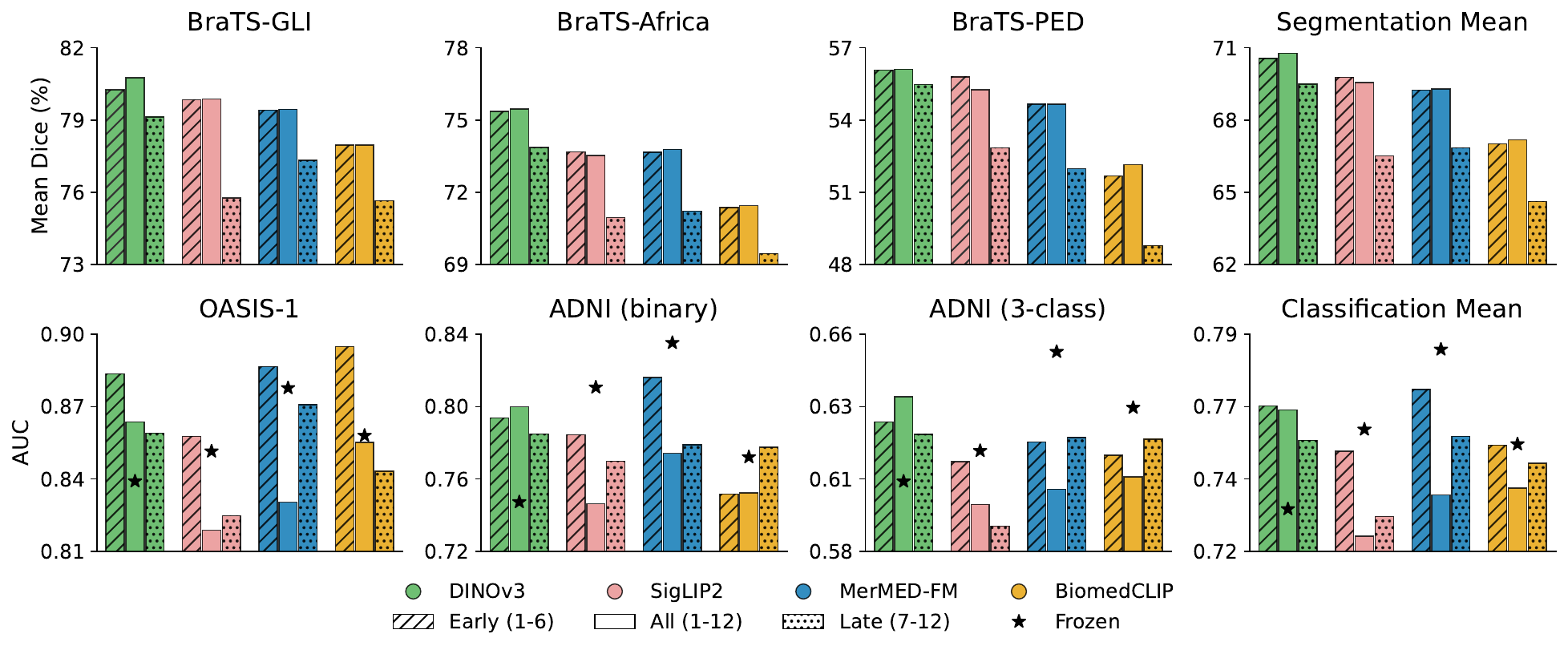}
    \caption{\textbf{Layer-selective LoRA performance across segmentation and classification datasets.} Each backbone is evaluated with three LoRA configurations: early (blocks 1--6, diagonal hatching), all (blocks 1--12, solid fill), and late (blocks 7--12, dotted hatching). Black stars indicate frozen baseline performance; when absent, it falls below the plotted range. Upper row: mean Dice (\%) on three segmentation datasets. Bottom row: AUC on three Alzheimer's disease classification benchmarks. The rightmost column shows each model's average across the three datasets in each task category.}
    \label{fig:layer_selective}
\end{figure*}

\subsection{Layer-Selective LoRA}\label{sec:results_layer_selective}

Figure~\ref{fig:layer_selective} summarises performance across the three LoRA configurations (\textit{early}, \textit{all}, \textit{late}) on all six datasets, with the frozen baseline marked as a black star where it falls within the plotted range. These experiments serve as a validation of the activation patching findings: if the early-layer recovery observed in Figure~\ref{fig:activation_patching} merely reflected signal accumulation rather than genuine causal sufficiency, restricting LoRA to early layers would substantially degrade performance.

In segmentation (upper row), early-layer LoRA matches or approaches all-layer performance for all four backbones across the three datasets, while late-layer LoRA consistently underperforms both. This validates the activation patching finding that early-to-mid layers are causally sufficient for segmentation. DINOv3 exhibits the smallest variation across configurations, consistent with its more uniform mechanistic profile (Figure~\ref{fig:mechanistic_panels}, panels a and b).

In classification (bottom row), the effect of layer selection is more pronounced. All-layer LoRA degrades classification relative to the frozen baseline for 3/4 backbones by up to ${\sim}$6\% AUC (Table~\ref{tab:crosstask_probe}); early-layer LoRA reduces this gap to ${\sim}$3\% or eliminates it entirely, and in some cases recovers or exceeds frozen performance. Late-layer LoRA, while occasionally outperforming all-layer adaptation, in the majority of the cases remains below early-layer performance and exhibits greater instability across datasets. This general pattern could suggest that the late-layer representational shift identified by activation patching may be responsible for the cross-task damage observed in Section~\ref{sec:results_class}: by not adapting those layers, early-layer LoRA preserves the representations that classification relies on. DINOv3, however, is a clear outlier in the classification results, in five out of the six tasks, the \textit{all} layer LoRA adaptation gives the best results. This is consistent with the classification results in Table \ref{tab:crosstask_probe}, where DINOv3 was the only backbone to show improved classification performance after LoRA adaptation.

\section{Discussion}

Our results show that four backbones with comparable segmentation performance diverge substantially in cross-task classification transfer. Mechanistic analysis traces this divergence to different representational shift profiles and reveals a dissociation between modification magnitude and causal importance: the largest attention divergence and weight changes concentrate in later layers (Figure~\ref{fig:mechanistic_panels}), yet activation patching identifies early-to-mid layers as causally sufficient for segmentation across all four backbones (Figure~\ref{fig:activation_patching}). The layer-selective LoRA experiments support this finding: restricting adaptation to early layers matches or approaches all-layer segmentation while mitigating the classification degradation that all-layer adaptation causes (Figure~\ref{fig:layer_selective}).

DINOv3 is the exception to this pattern. Although its later layers are also less causally important than its early ones, all-layer adaptation yields classification results similar to or better than early-layer adaptation, and DINOv3 is the only backbone whose classification performance improves after segmentation adaptation. Among the adapted models, it also shows the lowest JS divergence and weight-change ratios, the smallest variation across LoRA configurations, the highest mean Dice on all three segmentation datasets, and the best classification performance. This is compatible with the view that a smaller representational shift better preserves the representations that other tasks rely on. The relationship, however, is not deterministic: BiomedCLIP undergoes a substantial attention shift yet remains approximately neutral in classification on OASIS-1. The mechanistic tools are therefore best understood as diagnostics that reveal how a model adapts, rather than as direct predictors of downstream task performance.

In our experiments, medical-domain pre-training confers no consistent advantage for segmentation adaptation: after LoRA, DINOv3 achieves the highest and BiomedCLIP the lowest mean Dice on all three segmentation datasets, while MerMED-FM performs on par with SigLIP2. This is consistent with the hypothesis that representation quality scales with pre-training data diversity~\cite{liu2025biovfm}, though dataset scale and pre-training objective vary simultaneously, preventing causal attribution to any single factor. Notably, the frozen medical-domain models already attend to anatomically relevant regions more than their natural-domain counterparts (Figures~\ref{fig:mechanistic_panels}c and~\ref{fig:att_maps}), yet this initial advantage does not translate into superior adaptation performance, suggesting that where a model attends matters less than the quality of the representations it computes. In classification, the picture differs: frozen MerMED-FM achieves the highest AUC across all three settings, yet this advantage is entirely lost after segmentation adaptation, showing the largest classification drops among the four models. 

\vspace{0.5cm}

\noindent \textbf{Practical guidelines.} While our study is limited to four backbones and one imaging domain, the results suggest some considerations for selecting and adapting VFMs for medical imaging. First, backbone selection should not rely on segmentation metrics alone: evaluating on at least one additional task can reveal cross-task degradation invisible to Dice scores. Second, per-layer mechanistic profiles, particularly JS divergence and weight-change ratios, can serve as low-cost diagnostics before out-of-distribution data is available: in our experiments, the backbone with the most uniform and lowest divergence profile (DINOv3) consistently exhibited the best generalisation. Third, restricting LoRA to early-to-mid layers might be a conservative default when cross-task robustness matters, as it matched or approached full-layer segmentation for all four backbones while reducing classification degradation in 3/4 of them. Finally, we found no evidence that medical-domain pre-training provides an advantage over natural-domain pre-training for segmentation adaptation, though this comparison is confounded by dataset scale and should not be taken as a general conclusion.

\vspace{0.5cm}

\noindent \textbf{Limitations \& Future Work.} Our study is restricted to the brain MRI domain, to evaluating only four backbone models, to LoRA as the sole PEFT method, and to the ViT-B/16 architecture. In addition, whether the observed patterns generalise to other imaging modalities, adaptation strategies (e.g.\ adapters, prompt tuning), and larger model sizes (e.g.\ ViT-L, ViT-H) remains an open question. Segmentation experiments use three random seeds and classification probes use three cross-validation folds, which suffices to establish the reported trends but limits the precision of variance estimates. Future work would involve extending this framework to additional architectures, scales, imaging domains, and PEFT methods.

\section{Conclusion}

We have presented a mechanistic evaluation of four LoRA-adapted vision encoders for brain MRI, combining layer-wise interpretability tools with multi-task downstream assessment across six datasets. Our analysis reveals three key findings: (1)~segmentation convergence masks substantial divergence in classification transfer, with medical-domain pre-training conferring no consistent segmentation advantage in our setting; (2)~the most heavily modified layers are not the causally important ones, with early-to-mid layers consistently identified as sufficient for segmentation recovery; and (3)~restricting LoRA to these early layers matches or approaches full-layer segmentation while improving classification performance relative to all-layer adaptation in the three backbones whose cross-task performance degrades after adaptation.

Together, these findings support a practical protocol for adapting vision encoders in clinical settings: evaluate on multiple tasks to detect cross-task degradation invisible to segmentation metrics, use activation patching to identify causally relevant layers, and adapt preferentially those layers to balance task performance with representational robustness. While our study focuses on brain MRI and LoRA, the analytical framework is applicable to any attention-based encoder and PEFT method.

\hspace{2pt}

\textbf{Funding Statements.} 
C.D. is supported by Becas Santander Investigación Predoctoral and FPI-UAM grant. F.K. is supported by the UKRI Centre for Doctoral Training in AI for Healthcare (EP/S023283/1). A.K. is supported by a UKRI DTP award. A.G. and J.B. are supported by the Agencia Estatal de Investigación of Spain under Grant PID2025-171806OB-I00 and by the Regional Government of Madrid of Spain under Grant TEC-2024/COM-322. P.M.M. acknowledges generous personal support from the Edmond J. Safra Foundation and Lily Safra, an NIHR Senior Investigator Award, Rosalind Franklin Institute, and the UK Dementia Research Institute.  W.B. acknowledges the support of EPSRC CVD-Net Programme Grant (EP/Z531297/1) and BHF New Horizons Grant  (NH/F/23/70013). The authors gratefully acknowledge the HPC resources provided by NHR@FAU under the NHR projects b143dc and b180dc. NHR funding is provided by federal and Bavarian state authorities (HTA). NHR@FAU hardware is partially funded by the German Research Foundation (DFG~440719683). This work also received support from the ERC through project MIA-NORMAL (101083647). We further acknowledge the Isambard-AI AIRR, operated by the University of Bristol and funded by DSIT via UKRI and STFC [ST/AIRR/I-A-I/1023] \cite{mcintoshsmith2024isambardai}. \\

\textbf{Dataset Acknowledgements.} Data collection and sharing for the ADNI dataset used in this project were funded by the Alzheimer's Disease Neuroimaging Initiative (ADNI; NIH Grant U19AG024904). ADNI is funded by the National Institute on Aging and has previously received funding from the National Institute of Biomedical Imaging and Bioengineering, the Canadian Institutes of Health Research, and private sector contributions facilitated by the Foundation for the National Institutes of Health. The grantee organisation is the Northern California Institute for Research and Education. Data were provided in part by the Open Access Series of Imaging Studies (OASIS-1), led by principal investigators D. Marcus, R. Buckner, J. Csernansky, and J. Morris, and supported by grants P50 AG05681, P01 AG03991, P01 AG026276, R01 AG021910, P20 MH071616, and U24 RR021382.

\hspace{2pt}

\textbf{Data Availability}
The BraTS 2023 Adult Glioma, Sub-Saharan African Glioma, and Pediatric Glioma datasets used in this study are publicly available through the Synapse platform (\url{https://www.synapse.org/Synapse:syn51156910}). Similarly, OASIS-1 and ADNI are publicly available at \url{https://sites.wustl.edu/oasisbrains/home/oasis-1/} and \url{https://adni.loni.usc.edu/data-samples/adni-data/}, respectively.

\hspace{2pt}

\textbf{Code Availability}
Our code is publicly available at \url{https://github.com/fkekwick/Brain_Lora}. 

\hspace{2pt}

\textbf{CRediT Authorship Contribution Statement}
Cecilia Diana-Albelda: Conceptualization, Formal analysis, Investigation, Methodology, Validation, Visualization, Writing – original draft, Writing – review and editing.
Fiona Kekwick: Conceptualization, Formal analysis, Investigation, Methodology, Validation, Visualization, Writing – original draft, Writing – review and editing.
Raghav Mehta: Conceptualization, Methodology, Writing – review and editing.
Aleksandra Kulbaka: Writing – review and editing.
Paul M. Matthews: Funding acquisition, Supervision.
Jesus Bescos: Funding acquisition, Supervision, Writing – review and editing.
Álvaro García-Martín: Funding acquisition, Supervision, Writing – review and editing.
Wenjia Bai: Funding acquisition, Supervision, Writing – review and editing.
Bernhard Kainz: Conceptualization, Funding acquisition, Supervision, Writing – review and editing.


\hspace{2pt}

\textbf{Declaration of Competing Interests}
The authors declare that they have no known competing financial 
interests or personal relationships that could have appeared to 
influence the work reported in this paper.


\bibliographystyle{unsrt}

\bibliography{references}



\end{document}